%% file: main.tex
\documentclass[10pt,twocolumn,letterpaper]{article}

\usepackage{wacv}
\usepackage{times}
\usepackage{epsfig}
\usepackage{graphicx}
\usepackage{amsmath}
\usepackage{amssymb}

\usepackage{adjustbox}

\usepackage{mathtools}

\usepackage[ruled,vlined,linesnumbered]{algorithm2e}
\usepackage{algorithm2e}

\usepackage[pagebackref=true,breaklinks=true,letterpaper=true,colorlinks,bookmarks=false]{hyperref}
\allowdisplaybreaks

\wacvfinalcopy % *** Uncomment this line for the final submission

\def\wacvPaperID{545} % *** Enter the wacv Paper ID here

\ifwacvfinal\fi
\begin{document}

%%%%%%%%% TITLE
\title{DAC: Data-free Automatic Acceleration of Convolutional Networks
\vspace{-6mm}}
\author{Xin Li 
	\thanks{Xin Li and Shuai Zhang are equally contributed authors. This work is done while Xin Li is interning at Qualcomm.}
    \footnotemark[2] \footnotemark[3] , 
 	 Shuai Zhang\footnotemark[1] \footnotemark[2] , 
     Bolan Jiang\footnotemark[2] , 
     Yingyong Qi\footnotemark[2],
     Mooi Choo Chuah\footnotemark[3], \
     and Ning Bi\footnotemark[2] \\
\footnote{1} Qualcomm AI Research\\
\footnote{2} Department of Computer Science and Engineering, Lehigh University \\
{\tt\small xil915@lehigh.edu, shuazhan@qti.qualcomm.com, bjiang@qti.qualcomm.com}\\ 
{\tt\small yingyong@qti.qualcomm.com, chuah@cse.lehigh.edu, nbi@qti.qualcomm.com}\\
}

\maketitle
\ifwacvfinal\thispagestyle{empty}\fi

\input{abstract}
\input{introduction}
\input{priorwork}
\input{proposed_solution}

\input{experiments}
\input{conclusion}
\input{acknowledgements}

{\small
\bibliographystyle{ieee}
\bibliography{references}
}

%\begin{figure*}[h]
%	\centering
%	\includegraphics[width=\textwidth]{./images/appendix_openpose.pdf}
%	\caption{Visualized results of multi-person pose estimation on COCO dataset.}
%\end{figure*}
%
%\begin{figure*}[h]
%	\centering
%	\includegraphics[width=\textwidth]{./images/appendix_ssd.pdf}
%	\caption{Visualized results of object detection on PASCAL VOC2007 dataset. The results are generated by using the original model, DL+FL Rank7, and DL+FL Rank5 from left to right.}
%\end{figure*}

\end{document}

%% file: abstract.tex
%%%%%%%%% ABSTRACT
\begin{abstract}
Deploying a deep learning model on mobile/IoT devices is a challenging task. The difficulty lies in the trade-off between computation speed and accuracy. A complex deep learning model with high accuracy runs slowly on resource-limited devices, while a light-weight model that runs much faster loses accuracy. In this paper, we propose a novel decomposition method, namely DAC, that is capable of factorizing an ordinary convolutional layer into two layers with much fewer parameters. DAC computes the corresponding weights for the newly generated layers directly from the weights of the original convolutional layer. Thus, no training (or fine-tuning) or any data is needed. The experimental results show that DAC reduces a large number of floating-point operations (FLOPs) while maintaining high accuracy of a pre-trained model. If 2\% accuracy drop is acceptable, DAC saves 53\% FLOPs of VGG16 image classification model on ImageNet dataset, 29\% FLOPS of SSD300 object detection model on PASCAL VOC2007 dataset, and 46\% FLOPS of a multi-person pose estimation model on Microsoft COCO dataset. Compared to other existing decomposition methods, DAC achieves better performance. 

%DAC saves 45\% floating-point operations (FLOPS) of a VGG-like image classification model on CIFAR-10 dataset, 29\% FLOPS of SSD300 object detection model on PASCAL VOC2007 dataset, and 46\% FLOPS of a multi-person pose estimation model on Microsoft COCO dataset. 
\end{abstract}

%% file: introduction.tex
%%%%%%%%% BODY TEXT
\section{Introduction}

Deep learning techniques have been applied to many areas of artificial intelligence, which affects our daily lives.  For example, smart surveillance video systems that can detect and identify suspects help law enforcement personnel to maintain a safer living environment. Self-driving cars liberate drivers from steering wheels so that they can do more meaningful things, e.g., read business news. As technology for high-performance mobile or edge computing devices continues to improve, more and more deep learning models are deployed on these devices, e.g., face recognition systems are used on cell phones to unlock screens, etc. 

However, some of these AI tasks, e.g., voice recognition, requires internet access, which means the model is not entirely run on mobile/IoT devices. The major reason is that most of the deep learning models with high accuracy run too slowly on resource-limited devices. Many techniques to reduce the size of neural network models, e.g., model quantization of neural network models using fewer bits, have been proposed to facilitate their implementations on mobile chips \cite{shuai-LWB2018,shuai-Binaryrelax2018,shuai-BCGD2018,xu2018INQ-SJTU}. However, limited by current hardware structure and the tolerance for model accuracy drop, most of these quantization methods for real applications only focus on the 8-bit format. To further accelerate neural network models, it is more important to reduce computation complexity directly from the network architectures.   
Some research \cite{howard2017mobilenets, sandler2018mobilenetv2, Zhang2017shufflenet, Max2017L0pruning, li2017sbgar, li2018rehar} has been done to simplify these models before running them on mobile/IoT devices. Such research can be roughly categorized into two classes: 

\textbf{Designing new light-weight network architectures:} MobileNet proposed by Howard et al. in \cite{howard2017mobilenets,sandler2018mobilenetv2} is an excellent example. The model is based on a streamlined architecture that uses depthwise separable convolutions to build a light weight deep neural network. The model achieves good accuracy and runs fast on mobile devices. Similar with MobileNet, ShuffleNet \cite{Zhang2017shufflenet, shufflenet-V2-2018} is another type of light weight network architecture, based on depthwise separable layers for acceleration. 
However, these models require powerful servers and massive data to tune the weights. This is not a friendly solution to those who cannot access such resources. 

\textbf{Modifying an existing model to a slim version:} Another solution is to produce a slimmer version of an existing model. Unfortunately, the training data in some cases is exclusively available to the original designer of a model, which prevents other researchers from re-training the model after modification. Besides, it is costly and time-consuming to train a model from scratch. Thus, compared to designing new models and training them from scratch, accelerating an existing model based on its pretrained weights is a better solution. Network pruning and parameter decomposition are two common methods for this purpose. 
\textbf{Network pruning} is a practical tool for speeding up existing deep neural networks \cite{molchanov2016pruning}. He et al. propose a channel pruning method \cite{he2017channel} that utilizes LASSO regression to prune the number of the input channels in each convolutional layer.  Even though such network pruning scheme simplifies models, it still has some weaknesses. Network pruning is based on the statistical results of a set of samples. Thus: (1) it still requires data to discover which channel to prune, and (2) the accuracy of the model drops after pruning because the statistical results are not suitable for all data during testing. Louizos et al. incorporate $l_0$ relaxation \cite{Max2017L0pruning} into the training loss function to enforce compactness of network parameters. Thus, this $l_0$ pruning method should only be used during the training process. 
\textbf{Parameter decomposition} is another way to simplify an existing model. It is a layer-wise operation that decomposes a layer into one or multiple smaller layers, either having smaller kernel sizes or fewer channels. Although there will be more layers after being decomposed, the total number of weights and the computational complexity will be reduced. The decomposition methods only use the pre-trained weights of a layer, with the fact that most neural network models have much redundant parameters and can be largely simplified with low rank constraints. In this paper, we propose a new parameters decomposition method which does not require access to data or retraining.

The contributions of this paper are:

\textbf{1.} We propose a novel decomposition method that replaces standard convolutional layers in a pre-trained model with separable layers to significantly reduce the number of FLOPs. 
%In theory, the proposed method can be used on almost any convolutional layer in any model. 

\textbf{2.} The newly generated model maintains high accuracy without using any data and training process.
%We propose a novel non-training decomposition method, namely MobilizedNN, that replaces standard convolutional layers in a pre-trained model with separable layers to reduce a large number of parameters. 
%In theory, the proposed method can be used on almost any convolutional layer in any model. 

\textbf{3.} The experimental results on three computer vision application scenarios show that DAC maintains high accuracies even when a vast amount of FLOPs is trimmed. 

The rest of this paper is organized as follows. Some related works are summarized in section \ref{prior_work}. In section \ref{Solution}, we describe the architecture of DAC and our factorization method. The experimental results are reported in section \ref{experiments}, followed by the conclusion in Section \ref{conclusion}.

%% file: priorwork.tex
%-------------------------------------------------------------------
\section{Related Work }
\label{prior_work}

Much work has been done to do parameter decomposition. In this section, we will discuss some prior work that decomposes convolutional layers. To simplify the description, we assume the weight of the convolutional layer that we are going to decompose has a size of $(n \times k_w \times k_h \times c)$, where $n$ is the number of kernels, $k_w$ and $k_h$ are the spatial width and height of a kernel respectively, and $c$ is the number of channels of the input feature map.

First, Jaderberg et al. \cite{jaderberg2014Spatial_decomp} propose a spatial decomposition method. The method decomposes a convolutional layer with $(n \times k_w \times k_h \times c)$ kernel size into two layers. One has horizontal filters with $(c' \times k_w \times 1 \times c)$ kernel size and the other consists of vertical filters with $(n \times 1 \times k_h \times c')$ kernel size. In theory, this method indeed reduces parameters. However, running the decomposed model on a mobile device that has limited resources does not result in a significant speed up. This is due to the caching behavior of data. A feature map is horizontally (or vertically) loaded into a continuous block of memory. When we compute convolution using horizontal (vertical) filters, we access the memory sequentially. There is no impact on running time. However, if we compute the convolution using vertical (horizontal) filters, we cannot access memory sequentially any more which results in more cache misses and hence longer computation time. 

Then, Zhang et al. describe a channel decomposition method in \cite{Jian2016channel-decomp}. It decomposes a convolutional layer with $(n \times k_w \times k_h \times c)$ kernel size into a convolutional layer with fewer output channels and a pointwise convolutional layer. The newly generated convolutional layer has $(c' \times k_w \times k_h \times c)$ kernel size, and the pointwise convolutional layer has $(n \times 1 \times 1 \times c')$ kernel size. Notice that the first layer is also an ordinary convolutional layer, so it does not improve the situation fundamentally. 

Direct tensor decomposition methods including CP decomposition \cite{lebedev2014CP_decomp} and Tucker decomposition \cite{kim2015Tucker_decomp} are also applied to accelerate networks. After these tensor decompositions, one convolution layer will be factorized into 3 or 4 small layers with a bottleneck structure, opposite with \cite{sandler2018mobilenetv2} architecture. One big disadvantage of these tensor decomposition methods is that the depth of network architecture is tripled (3x) compared to the original model, thus it increases the memory access cost (MAC) and largely offset the gains from the reduction of FLOPs, as claimed in \cite{shufflenet-V2-2018}. 

There are also many network decomposition works using low rank constraints in training process or solving layer-wise regression problem with data samples \cite{wen2017Filter-LowRank,alvarez2017Proximal-NIPs}. But all these methods require the access of sufficient data from training/test domain. 

Our research focus is based on the real application scenario with limited access of data. In this paper, we propose a novel data-free convolutional layers decomposition method and compare its performance to two most related works \cite{Jian2016channel-decomp,jaderberg2014Spatial_decomp}. (After this paper was accepted, we found Guo et al. proposed a similar solution in \cite{guo2018network}. These two works are independent and concurrent.)

%% file: proposed_solution.tex
%-------------------------------------------------------------------
\section{Proposed Solution}
\label{Solution}

\begin{figure*}[h]
	\centering
	\includegraphics[width=0.85\textwidth]{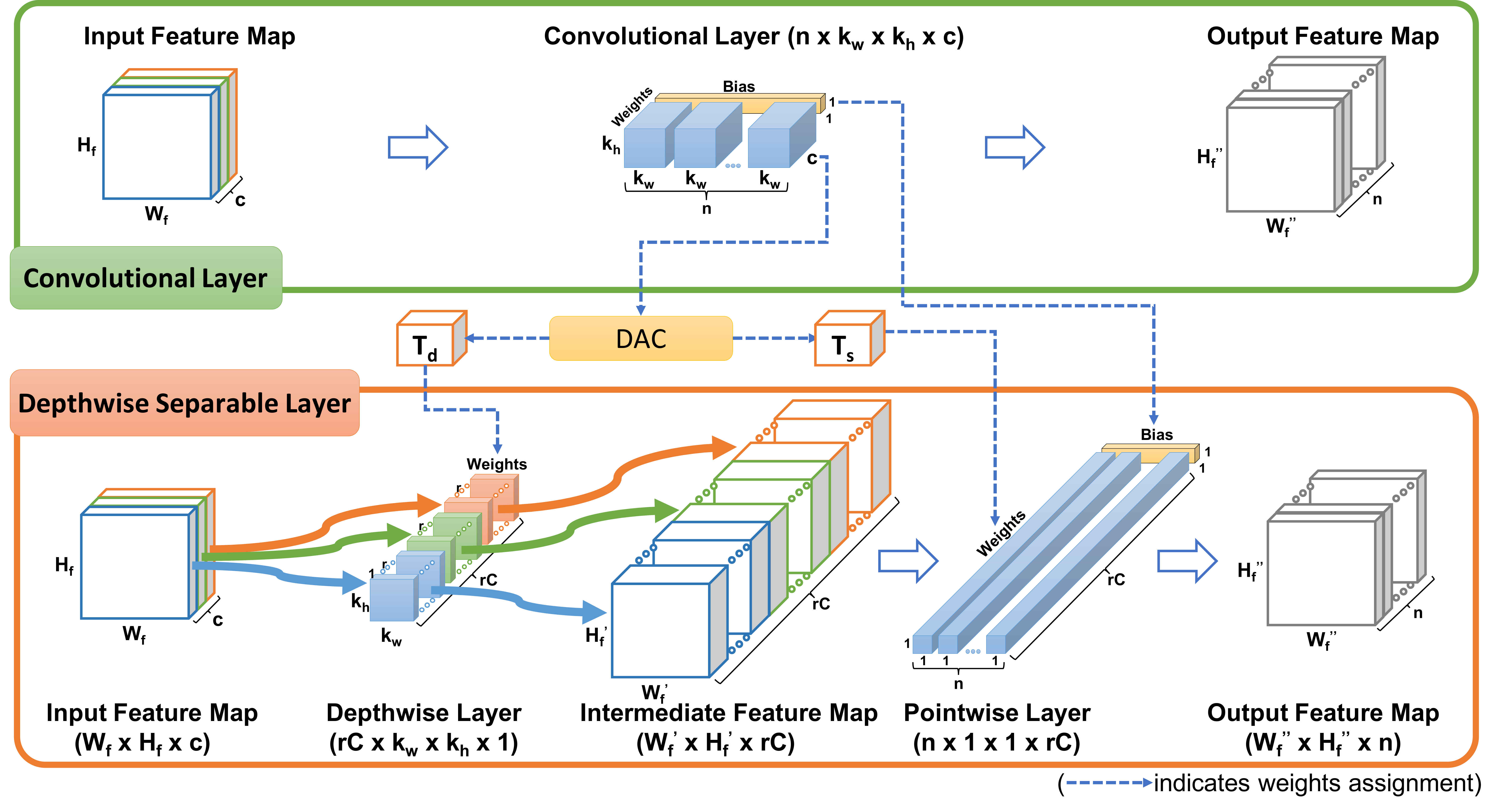}
	\caption{The architecture of our proposed DAC. An input feature map consists of $c$ channels (in this figure, $c = 3$) is marked with different colors. In ``Depthwise Layer'', kernels are only applied on the channel with the same color. Thus, each channel is processed by $r$ kernels.}
	\label{fig:dac_architecture}
\end{figure*}

The intuition of our proposed scheme is that the depthwise  + pointwise combination runs efficiently on mobile devices has already been proven by MobileNet \cite{howard2017mobilenets}. It will be useful if we can convert an ordinary convolutional layer into such a structure and compute their weights from the original layer directly. The feasibility of decomposing the weights of a convolutional layer has been mathematically proved by Zhang et al. \cite{Jian2016channel-decomp}.

\subsection{Convolutional Layer Factorization}
In this section, we propose a novel factorization method for convolutional layers. Figure \ref{fig:dac_architecture} shows the details of our scheme. An ordinary convolutional layer with the shape of $(n \times k_w \times k_h \times c)$ is decomposed into two layers. One is a depthwise layer with the shape of $(rC \times k_w \times k_h \times 1)$, and the other is a pointwise layer with the shape of $(n \times 1 \times 1 \times rC)$, where $rC = r * c$ and $r$ is a factor used to balance the trade-off between model compression ratio and accuracy drop. There is no bias in the depthwise layer, and the bias vector in the original layer is assigned to the pointwise layer.

Even though our scheme is inspired by MobileNet, it is worth highlighting the differences between MobileNet and DAC. DAC has no non-linear layers (batch normalization layers and activation layers) between the depthwise and the pointwise layers. The absence of non-linear layers makes DAC quantization friendly and hence suitable for further hardware acceleration, which Sheng et al.  \cite{NPU2018Quant_mobile} have already experimentally verified. 

\subsection{Weights Decomposition}
Once a convolutional layer is factorized, we want to compute weights for the newly generated layers (a depthwise and a pointwise layer) from the original weights directly. We assume $T$ is the trained weights of the original convolutional layer, and its shape is $(n \times k_w \times k_h \times c)$. We denote $Td \in D := \mathbb{R}^{rC \times k_w \times k_h \times 1}$ as the weights of the depthwise layer and $Ts \in S := \mathbb{R}^{n \times 1 \times 1 \times rC}$ as the weights of the pointwise layer. Then, the objective function of factorizing a convolutional layer is:

\begin{equation}
 \min \limits_{Td \in D, Ts \in S}  \|  T - Ts * Td \|_F^2,
\end{equation}
where operator $*$ is the combination of convolution operations of the depthwise and the pointwise layer, and $\| \|_F$ is the Frobenius norm 
for tensor/matrix. Thus 
\begin{align*}
	& \min \limits_{Td \in D, Ts \in S}  \|T - Ts * Td \|_F^2 \\
    & = \min \limits_{Td \in D, Ts \in S} 
    	\sum \limits_{i=1}^{C} \|T_i - Ts_i * Td_i \|_F^2  \\
    & = \sum \limits_{i=1}^{C} 
    	\min \limits_{Td_i, Ts_i} \|T_i - Ts_i * Td_i \|_F^2 \\
    & =  \sum \limits_{i=1}^{C} 
    	\min \limits_{S_i, D_i} \|M_i - S_i D_i \|_F^2.
 \vspace{-4mm}
\end{align*}
Here matrices $M_i$, $S_i$ and $D_i$ are transformed from tensors $T_i$, 
$Ts_i$ and $Td_i$ respectively. 

According to the SVD theory, the solution of minimization problem 
$\min \limits_{S_i, D_i}   \| M_i - S_i D_i \|_F^2$ 
is the singular matrices with rank $r$, where the top $r$ singular 
values can be merged into either $S_i$ or $D_i$. 
Also, Frobenius norm $\| \|_F$ can be defined as $\| \|_{2,2}$ 
induced by $L_2$ vector norm, so the above DAC minimization 
objective function can be considered as 
\begin{align*}
& \min \limits_{Td \in D, Ts \in S}  \|  T - Ts * Td \|_F^2  \\
& = \min \limits_{Td \in D, Ts \in S} \sup \limits_{\|F\|_2 \neq 0} 
 	\dfrac{\|(T - Ts * Td)F \|_2}{\|F\|_2},
\end{align*}
where $F$ is the input feature maps and $\|F\|_2$ is the vector $L_2$ norm. 
In this formula, it minimizes the output feature maps with approximation 
error measured in Euclidean space and the constraint of the decomposition `rank'  $r$ (the factor used to balance the trade-off between model compression ratio and accuracy drop). The process of weights decomposition is described in Algorithm \ref{algorithm:DAC}. 

\begin{algorithm}[h]
	\footnotesize
	\caption{DAC Weights Decomposition}
	\label{algorithm:DAC}
	\DontPrintSemicolon
	\SetKwInOut{Input}{Input}\SetKwInOut{Output}{Output}
	\Input{
		Weights of a convolutional layer: $T \in \mathbb{R}^{n \times k_w \times k_h \times c}$;\\
		Decomposition Rank: $r$.
		}
	\Output{
		Weights of the depthwise layer: $Td \in \mathbb{R}^{rC \times k_w \times k_h \times 1}$; \\
		Weights of the pointwise layer: $Ts \in \mathbb{R}^{n \times 1 \times 1 \times rC}$ 
		}
	\Begin{
		$list\_d \in \mathbb{R}^{c \times r \times k_w \times k_h \times 1} \leftarrow \emptyset$ \\
		$list\_s \in \mathbb{R}^{n \times 1 \times 1 \times r \times c} \leftarrow \emptyset$ \\		
		\For{$i \in c$}{
		$T_i \leftarrow T[:,:,:,i] \in \mathbb{R}^{n \times k_w \times k_h}$ \\
		$M_i \leftarrow Reshape(Ti, (n, k_w \times k_h)) \in \mathbb{R}^{n \times k_wk_h}$ \\
		$D_i, S_i \leftarrow Decompose(M_i, r)$ \\
		$list\_d[i,:,:,:,:] \leftarrow D_i \in \mathbb{R}^{r \times k_w \times k_h \times 1}$ \\
		$list\_s[:,:,:,:,i] \leftarrow S_i \in \mathbb{R}^{n \times 1 \times 1 \times r}$ \\
 	}
 	$Td \leftarrow Reshape(list\_d, (r \times c, k_w, k_h, 1 )) $ \\
 	$Ts \leftarrow Reshape(list\_s, (n, 1, 1, r \times c)) $ \\ 	
	}
	\textbf{function} Decompose(M, r) \\
	\Begin{
		%\footnotesize
		$U, Sigma, V \leftarrow SVD(M)$ \\
		$Ur \leftarrow U[:, :r] \in \mathbb{R}^{n \times r}$ \\
		$Vr \leftarrow V[:r, :] \in \mathbb{R}^{r \times k_wk_h}$ \\
		$Sr \leftarrow Sigma[:r, :r] \in \mathbb{R}^{r \times r}$ \\		
		$D \leftarrow Reshape(Vr, (r, k_w, k_h, 1))$ \\
		$S \leftarrow Ur \, Sr$ \\
		$S \leftarrow Reshape(S, (n, 1, 1, r))$ \\
		\Return{$D, S$}
	}
\end{algorithm}

\subsection{Computation Reduction}
We consider the original convolutional layer with $(n \times k_w \times k_h \times c)$ kernel size takes a $(W_f \times H_f \times c)$ feature map $F$ as an input and produces a $(W_f \times H_f \times n)$ feature map $G$, where $W_f$ and $H_f$ are the spatial width and height of the feature maps. Here, we assume the output feature map has the same spatial size as the input for simplification. Then, the computation cost of the convolutional layer is: $W_f \times H_f \times c \times k_w \times k_h \times n.$

The computation cost depends on the number of input channels $c$, the number of output channels $n$, the kernel size $k_w \times k_h$ and the input features map size $W_f \times H_f$. 
%DAC breaks up the interaction among them. 
After decomposition, the newly generated depthwise and pointwise layer in total have the cost of $W_f \times H_f \times k_w \times k_h \times rC + W_f \times H_f \times rC \times n,$ where $rC = r * c$ and the reduction in computation is
\begin{align*}
& \dfrac{W_f \times H_f \times k_w \times k_h \times rC + W_f 
    \times H_f \times rC \times n}{W_f \times H_f \times c \times k_w \times k_h \times n}  \\
&= \frac{r}{n} + \frac{r}{k_wk_h}
\end{align*}

%% file: experiments.tex
%-------------------------------------------------------------------
\section{Experimental Results} 
\label{experiments}
To prove the universality of our proposed scheme, we apply DAC to three major application scenarios in the field of Computer Vision: (1) Image Classification, (2) Object Detection, and (3) Multi-person Pose Estimation. We implement our scheme using Python and Keras Library \cite{chollet2015keras} with Tensorflow backend \cite{abadi2016tensorflow}.

\subsection{Datasets}
Four datasets are used in this paper: 

\textbf{CIFAR-10 dataset:}  The CIFAR-10 dataset \cite{krizhevsky2009learning} consists of 50,000 training images and 10,000 test images in 10 categories. It is a small dataset, from which we can quickly get results after tuning parameters.  Thus, we use it for ablation study to get some insights about DAC, e.g., the impacts of using different ranks or decomposing different layers. 

\textbf{ImageNet dataset:} The ImageNet dataset \cite{russakovsky2015imagenet} has 50,000 ILSVRC validation images in 1,000 object categories. We use this ILSVRC validation subset to evaluate the performance of DAC in the task of image classification. 

\textbf{Pascal VOC2007 dataset:} For object detection task, Pascal VOC2007 dataset \cite{pascal-voc-2007} is used. It consists of 4,952 testing images for object detection. The bounding box and label of each object from twenty target classes have been annotated. Each image has one or multiple objects. 

\textbf{Microsoft COCO dataset:} The Microsoft COCO dataset \cite{lin2014microsoft} is used to evaluate the performance of DAC in the task of multi-person pose estimation. We use the COCO 2017 keypoints subset which consists of 5,000 validation images and ~40K testing images. 

\subsection{Ablation Study}
\begin{figure}[h]
	\vspace{-2mm}
	\centering
	\includegraphics[width=0.48\textwidth]{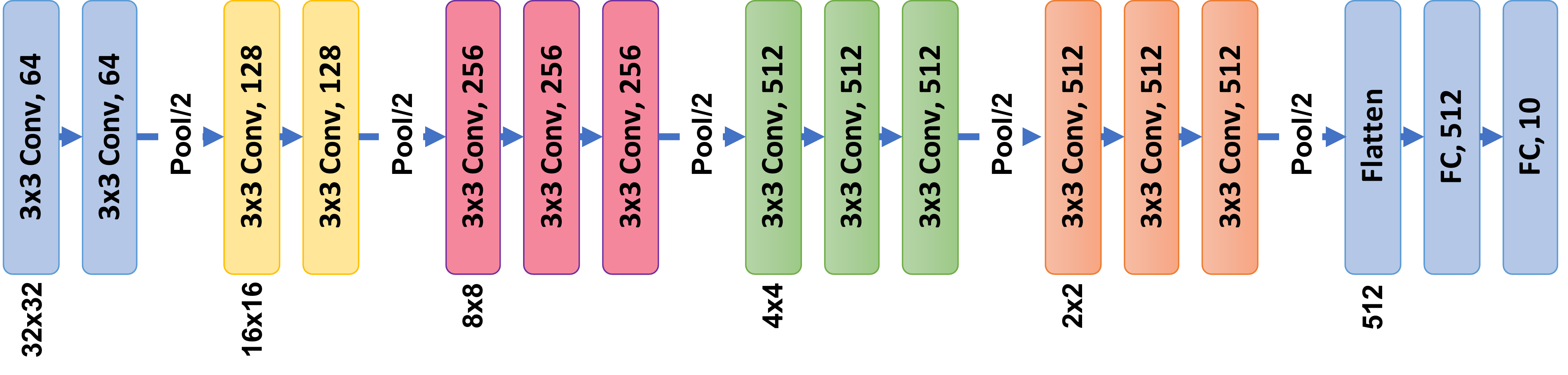}
	\caption{The architecture of the CIFAR-VGG.}
	\label{fig:cifar_vgg}
\end{figure}

Here, we use a pre-trained CIFAR-VGG model \footnote{https://github.com/geifmany/cifar-vgg}, a simple Convolutional Neural Network, on the CIFAR-10 dataset as our original model. Figure \ref{fig:cifar_vgg} shows the architecture of the CIFAR-VGG. In total, the CIFAR-VGG model has 13 convolutional layers. The original model (trained on CIFAR-10 training subset) achieves 93.6\% on CIFAR-10 testing subset.  

First, we decompose a single convolutional layer to explore the impact of decomposing different layers. Table \ref{table:cifarvgg_singlelayer} shows the details of testing accuracy when applying varying ranks (rank 1 to rank 5) decomposition on different layers of CIFAR-VGG model. Each time, we only modify one layer. All results are collected using decomposed weights directly (no access to data or any training process). 

\begin{table}[h]
  \centering  
  \caption{Testing Accuracy on CIFAR-10 dataset when decomposing different layers of CIFAR-VGG model using variant ranks.}
  \begin{adjustbox}{width=0.5\textwidth}
  \begin{tabular}{|l|c c c c c|} 
    \cline{2-6} 
  \multicolumn{1}{c|}{}& \multicolumn{5}{|c|}{\textbf{Accuracy (\%)}}  \\
  \hline
  \multicolumn{1}{|c|}{\textbf{Original Model}} & \multicolumn{5}{|c|}{93.6} \\
    \hline
    \hline
    \textbf{Decomposed Layer} & \textbf{Rank1} & \textbf{Rank2} & \textbf{Rank3} & \textbf{Rank4} & \textbf{Rank5} \\
    \hline    
    \textbf{conv2d\_1} & 18.6 & 76.4 & 86.7 & 91.9 & 92.8 \\
    \textbf{conv2d\_2} & 39.1 & 86.5 & 91.6 & 92.7 & 93.2 \\
    \textbf{conv2d\_3} & 54.0 & 87.8 & 92.6 & 93.2 & 93.4 \\
    \textbf{conv2d\_4} & 31.4 & 83.7 & 91.8 & 92.8 & 93.2 \\
    \textbf{conv2d\_5} & 80.1 & 90.2 & 92.6 & 93.1 & 93.8 \\
    \textbf{conv2d\_6} & 84.3 & 90.9 & 92.8 & 93.3 & 93.4 \\
    \textbf{conv2d\_7} & 66.0 & 89.5 & 92.6 & 93.0 & 93.3 \\
    \textbf{conv2d\_8} & 83.2 & 91.1 & 92.6 & 93.0 & 93.2 \\
    \textbf{conv2d\_9} & 91.2 & 93.1 & 93.3 & 93.4 & 93.5 \\
    \textbf{conv2d\_10} & 91.7 & 93.2 & 93.3 & 93.3 & 93.4 \\
    \textbf{conv2d\_11} & 93.1 & 93.4 & 93.3 & 93.4 & 93.4 \\
    \textbf{conv2d\_12} & 93.3 & 93.4 & 93.4 & 93.4 & 93.5 \\
    \textbf{conv2d\_13} & 92.9 & 93.4 & 93.4 & 93.4 & 93.5 \\
    \hline
  \end{tabular}
  \end{adjustbox}
   \label{table:cifarvgg_singlelayer}
   \vspace{-3mm}
\end{table} 

From Table \ref{table:cifarvgg_singlelayer}, we gain two insights: (a) Decomposing first few layers of a model causes large drops in accuracy (75\% drop when rank 1 decomposition is applied on layer conv2d\_1), while decomposing last few layers has a smaller impact on the accuracy (less than 1\% drop when rank 1 decomposition is applied on layer conv2d\_13). (b) Decomposing a layer using a larger rank helps to maintain the accuracy. This can be observed by comparing different columns in the same row. These two insights are consistent with our intuition. (a) Decomposing a layer generates tiny errors. If such errors occur at the beginning of a model, the errors will accumulate to bigger errors at the final prediction. (b) Compared to smaller ranks, larger ranks generate more parameters in the depthwise layers. Thus, the newly generated layers have more possibility of replicating the performance of the original layer. 

\begin{figure}[h]
	\centering
	\includegraphics[width=0.49\textwidth]{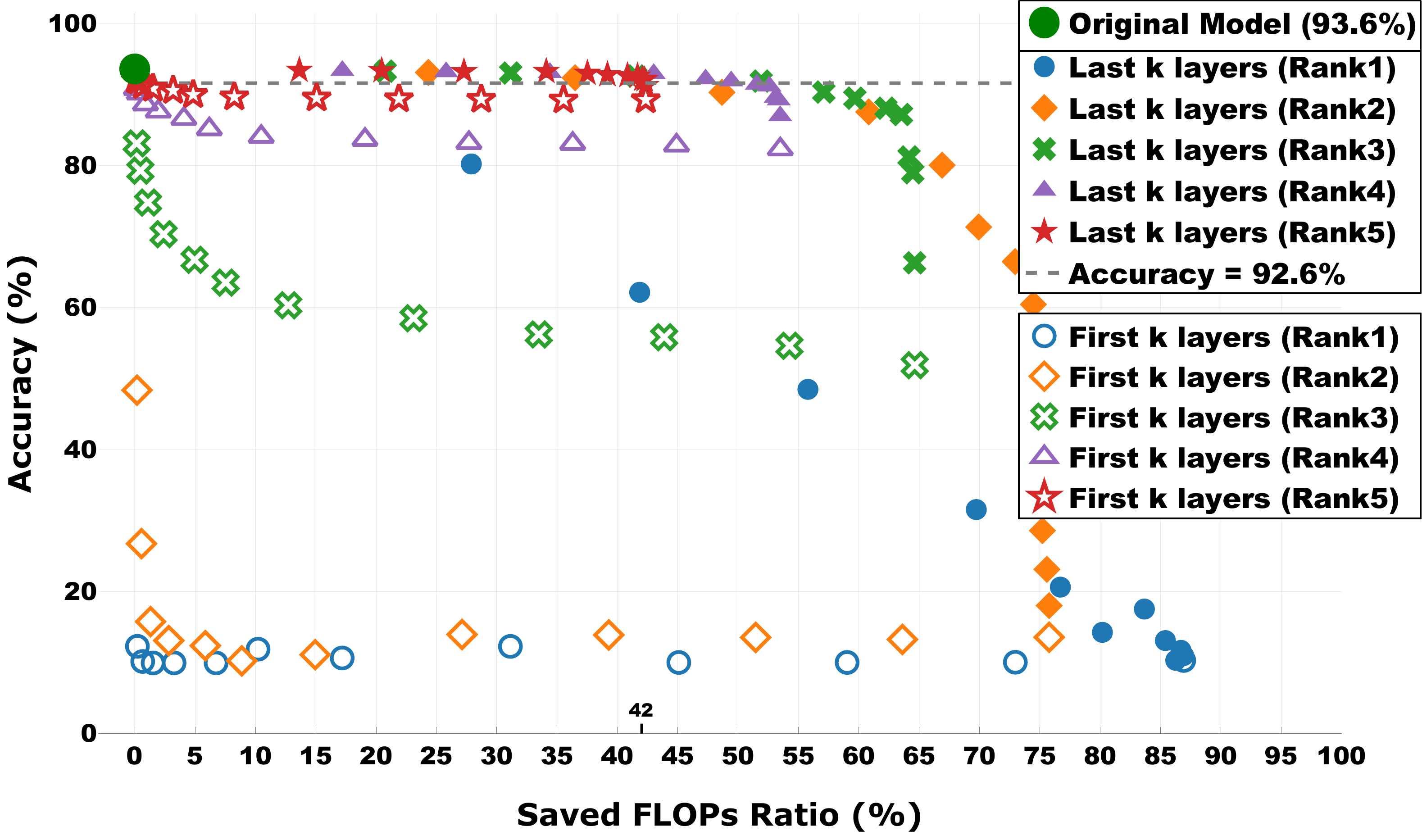}
	%\caption{Classification accuracy on the CIFAR-10 dataset. Orange-triangle points indicate the highest accuracy at a specific ratio of saved FLOPs. Red-star points are the cases where all modified layers are decomposed using Rank 5. In addition, we annotated the modified layers of 4 red points, e.g. (12-13) indicates that layer ``conv2d\_13'' and ``conv2d\_12'' are decomposed with Rank 5.}
	\caption{Classification accuracy on the CIFAR-10 dataset. Each curve has 12 points that correspond to different numbers of decomposed layers (2 to 13 layers from left to right). Solid spots indicate the cases that last few layers are decomposed (layer ``conv2d\_13'' included). Open spots are the cases that first few layers are decomposed (``conv2d\_1'' layer included).}
	\label{fig:cifar10_acc}
	\vspace{-5mm}
\end{figure}

%Next, we explore the performance of DAC when multiple convolutional layers are decomposed. Based on the first insight from above, we decompose the last few convolutional layers with varying ranks (Rank 1 to Rank 5). The experimental results are reported in Figure \ref{fig:cifar10_acc}. First, one can quickly notice that most decomposition cases achieve high accuracies (higher than 91.6\% or 2\% drop). Second, after saving 42\% FLOPs, DAC still achieves 92.7\% accuracy (drops less than 1\%). Both of these prove that our proposed DAC has the capability of maintaining accuracy when the number of FLOPs is substantially reduced. 
Next, we explore the performance of DAC when multiple convolutional layers are decomposed. We decompose the model with two opposite directions: (1) from the last layer to the first one, and (2) from the first layer to the last one. To simply the experiment, we use the same rank to decompose all chosen layers. The experimental results are reported in Figure \ref{fig:cifar10_acc}. First, one can quickly notice that most decomposition cases (solid points) achieve high accuracies (higher than 91.6\% or 2\% drop). Second, after saving 42\% FLOPs, DAC still achieves 92.7\% accuracy (drops less than 1\%). Both of these prove that our proposed DAC has the capability of maintaining accuracy when the number of FLOPs is substantially reduced. 

Besides, in Figure \ref{fig:cifar10_acc}, red-star points (Rank 5) achieve high accuracies. If we compare the solid (open) red-star marks to other solid (open) marks, we can notice that the above insights also hold in the case of decomposing multiple convolutional layers. Ten (eight) out of twelve Rank 5 decomposition cases (solid red-star spots) drop accuracy by less than 2\% (1\%). The worst solid red-star case that achieves 91.2\% (accuracy drops 2.4\%) is caused by the decomposition of the first layers of the model (first insight discussed above). It is worth highlighting that these decomposed models that maintain high accuracies are generated by DAC without accessing data or training process.

\subsection{Image Classification}
For the task of image classification, we use the VGG16 model proposed by Simonyan et al. in \cite{simonyan2015very}. It includes 12 (3x3) convolutional layers. We downloaded a model \footnote{https://github.com/fchollet/deep-learning-models/releases/download/v0.1/vgg16\_weights\_tf\_dim\_ordering\_tf\_kernels.h5} pre-trained on ImageNet dataset. All convolutional layers but the first one are decomposed considering the first insight we got in our ablation study. 

\begin{table}[h]
\vspace{-2mm}
  \centering  
  \caption{Top-5(Top-1) Validation Accuracy on ImageNet dataset}
  \begin{adjustbox}{width=0.45\textwidth}
  \begin{tabular}{|l|c c c|} 
    \cline{2-4} 
  \multicolumn{1}{c|}{}& \multicolumn{3}{|c|}{\textbf{Top-5(Top-1) Accuracy (\%)}}  \\
  \hline
  \multicolumn{1}{|c|}{\textbf{VGG16 \cite{simonyan2015very} (Baseline) }} & \multicolumn{3}{|c|}{88.9(69.2)} \\
    \hline
    \hline
    \textbf{Method} & \textbf{Saved 40\%} & \textbf{Saved 50\%} & \textbf{Saved 60\%} \\
    \hline    
	\textbf{Channel Decomp. \cite{Jian2016channel-decomp}} & 86.5(65.6) & 74.4(48.7) & 43.3(20.8) \\
    \textbf{Spatial Decomp. \cite{jaderberg2014Spatial_decomp}} & 88.6(68.5) & 86.3(65.0) & 78.0(52.5) \\
    \textbf{DAC (Ours)} & \textbf{88.6(68.5)} & \textbf{87.5(66.8)} & \textbf{84.7(62.5)}\\
    \hline
  \end{tabular}
  \end{adjustbox}
  \label{table:imagenet_acc}
  \vspace{-3mm}
\end{table} 

Here we compare our approach with two schemes, namely, the Filter Reconstruction Optimization proposed by Jaderberg et al. in \cite{jaderberg2014Spatial_decomp} (Spatial Decomp. in Table \ref{table:imagenet_acc}) and the Channel Decomposition method proposed by Zhang et al. in \cite{Jian2016channel-decomp} (Channel Decomp. in Table \ref{table:imagenet_acc}). Spatial Decomposition  is the one that does not need data and training like DAC as we discussed in Section \ref{prior_work}. Although the Channel Decomposition requires some data, we can still use the method as a filter reconstruction without accessing any data and training process. We implemented these two algorithms ourselves. For fair comparison, we choose appropriate parameters for Channel Decomposition and Spatial Decomposition, so that all schemes save roughly same FLOPs. Given a rank $r$ of DAC, the number of filters $c'_c$ in the first newly generated layer in Channel Decomposition can be computed using:

\begin{equation}
	\vspace{-1mm}
	c'_c = r * \frac{c(n+k_hk_w)}{ck_hk_w+n}
	\label{equation:rank_channel}
	\vspace{-1mm}
\end{equation}
and for Spatial Decomposition,  the number of filters $c'_s$ in the first newly generated layer is 

\begin{equation}
	\vspace{-2mm}
	c'_s = r * \frac{c(n+k_hk_w)}{ck_w+nk_h}
	\label{equation:rank_spatial}
\end{equation}
where $n$ is the number of kernels in original convolutional layer, $k_w$ and $k_h$ are the spatial width and height of a kernel respectively, and $c$ is the number of channels of the input feature map. 

Table \ref{table:imagenet_acc} shows the accuracy of the model (after saving 40\%, 50\%, and 60\% FLOPs respectively) on ImageNet validation set. First, DAC maintains high accuracy on both Top-1 and Top-5 accuracy even when a significant amount of FLOPs are reduced. Second, compared to the Channel Decomposition and Spatial Decomposition, DAC performs much better. Especially when we saved 60\% FLOPs, DAC achieves 41.4\% higher accuracy than Channel Decomposition and 6.7\% higher accuracy than Spatial Decomposition. 

\subsection{Multi-person Pose Estimation}
\begin{figure}[h]
	\centering
	\vspace{-3mm}
	\includegraphics[width=0.4\textwidth]{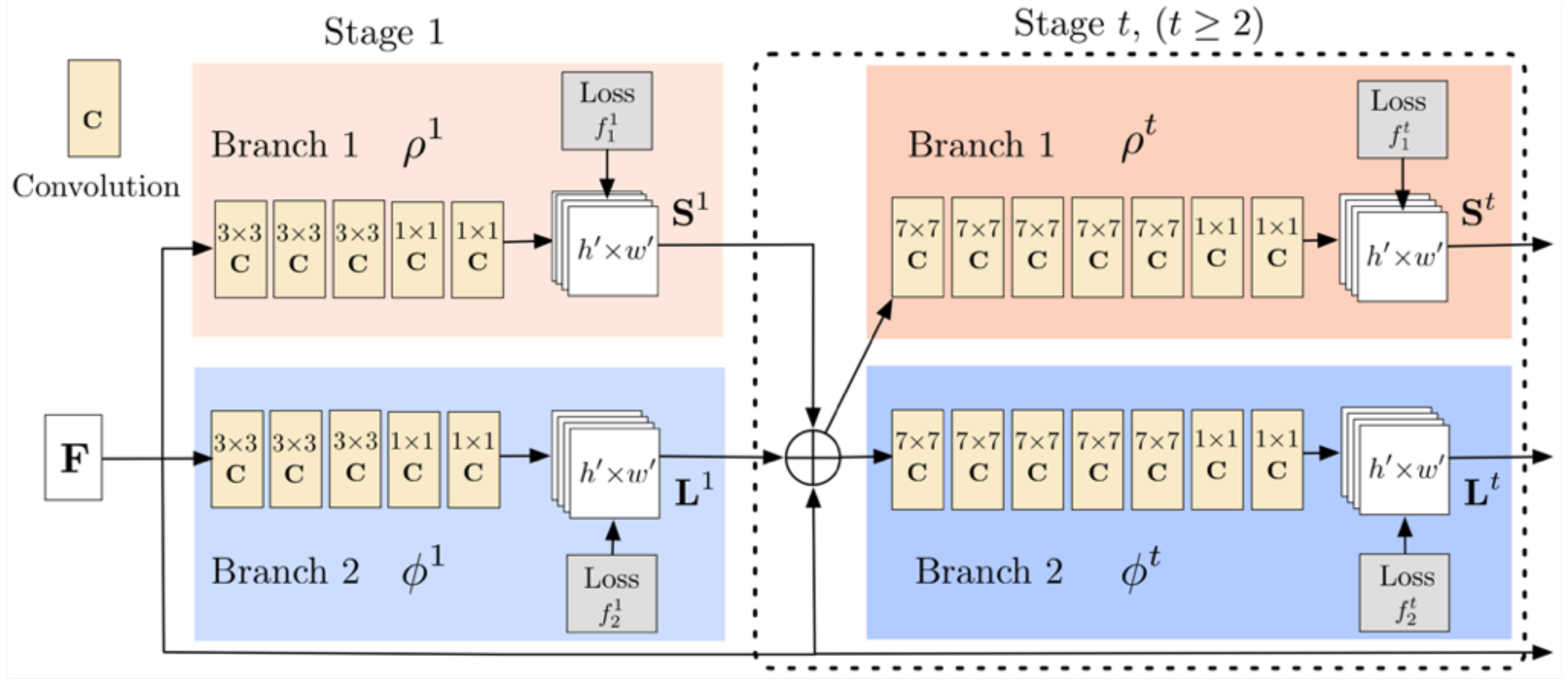}
	\vspace{-1mm}
	\caption{The model architecture figure extracted from \cite{cao2017realtime}.}
	\label{fig:openpose}
	\vspace{-2mm}
\end{figure}

For the task of multi-person pose estimation, we use the scheme proposed by Cao et al. \cite{cao2017realtime}. Figure \ref{fig:openpose} is the architecture extracted from their paper. After generating the feature map $F$ by a convolutional network (initialized by the first 10 layers of VGG-19 \cite{simonyan2015very} and fine-tuned), the model is split into two branches: the top branch predicts the confidence maps, and the bottom branch predicts the affinity fields. 

We download an implementation of Cao's model \footnote{https://github.com/anatolix/keras\_Realtime\_Multi-Person\_Pose\_Estimation} that was pre-trained on Microsoft COCO dataset as our original model. It achieves 57.9\% average precision (AP) on the validation subset of 2017 COCO keypoints challenge. This model consists of six stages, which means $t \in \lbrace 2, 3, 4, 5, 6 \rbrace$ in Figure \ref{fig:openpose}. Thus, the first stage (Stage 1) has 6 convolutional layers (3x3 kernel size), and each of the following stage (Stage 2 to Stage 6) includes 10 convolutional layers (7x7 kernel size). Based on the above two insights, we decompose the model from the bottom to the top with variant ranks (from Rank20 to Rank3). Because the full rank of a (3x3) convolutional kernel (in Stage 1) is 9, so we set the maximum rank used to decompose these (3x3) convolutional layers equals to 5 for a large compression ratio. 

\begin{figure}[h]
	\centering
	\vspace{-3mm}
	\includegraphics[width=0.45\textwidth]{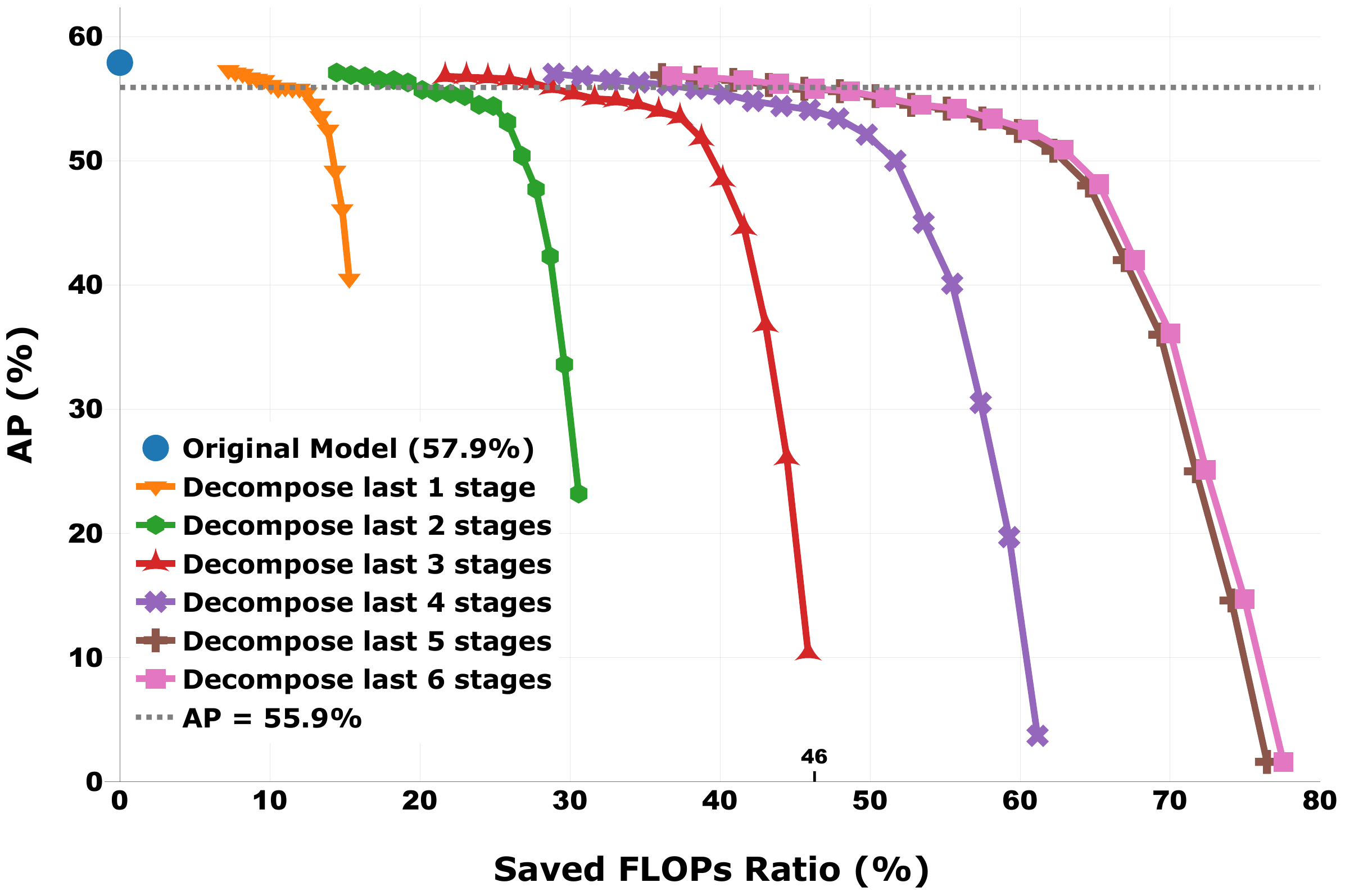}
	\caption{Results on the Microsoft COCO dataset. Each curve has 18 points that correspond to different ranks (Rank20 to Rank3 from left to right). }
	\label{fig:openpose_acc}
	\vspace{-3mm}
\end{figure}

\begin{figure*}[h]
	\centering
	\includegraphics[width=0.88\textwidth]{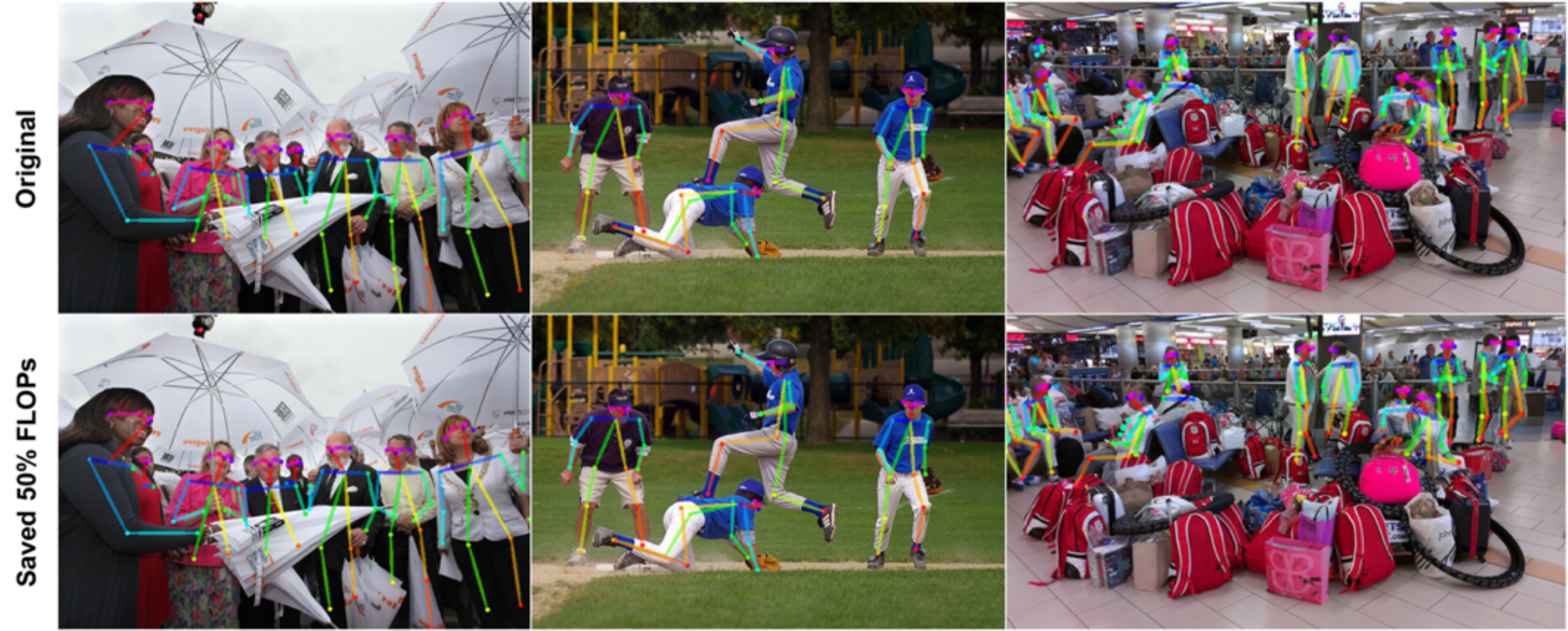}
	\caption{Visualized results on COCO dataset. The first row shows the results generated using the original weights, while the second row shows the results created using the model that saves 50\% FLOPs. }
	\label{fig:openpose_vis}
	\vspace{-4mm}
\end{figure*}

Figure \ref{fig:openpose_acc} shows the experimental results. First, it is obvious that in the task of person pose estimation, the DAC also maintains high accuracy without any retraining when large amounts of FLOPs are saved. Our proposed DAC saves up to 46\% FLOPs when 2\% AP drop is allowed. Second, for each curve, the AP decreases with  decreasing decomposition rank. This observation is consistent with the above second insight. Then, we notice that ``Decompose last 6 stages'' achieves similar results (similar saved ratios and APs) as ``Decompose last 5 stages'' does. This can be explained as follows: the ``Decompose last 6 stages'' includes Stage 1 in which all decomposed convolutional layers (6 layers) have (3x3) kernel size. Comparing to a convolutional layer with (7x7) kernel size, these layers have much fewer parameters, so decomposing them does not contribute much. 
%In addition, we set the maximum rank used to decompose these (3x3) convolutional layers equals to 5, so first 16 spots on the ``Decompose last 6 stages'' curve have the same decomposition rank applied on the layers in Stage 1. 

\begin{table}[h]
  \vspace{-2mm}
  \centering  
  \caption{Results on the COCO 2017 keypoint challenge}
  \begin{adjustbox}{width=0.45\textwidth}
  \begin{tabular}{|l|c c c|} 
    \cline{2-4} 
  \multicolumn{1}{c|}{}& \multicolumn{3}{|c|}{\textbf{Mean Average Precision (\%)}}  \\
  \hline
  \multicolumn{1}{|c|}{\textbf{Openpose \cite{cao2017realtime} (Original) }} & \multicolumn{3}{|c|}{57.9} \\
    \hline
    \hline
    \textbf{Method} & \textbf{Saved 40\%} & \textbf{Saved 50\%} & \textbf{Saved 60\%} \\
    \hline    
	\textbf{Channel Decomp. \cite{Jian2016channel-decomp}} & 25.9 & 5.0 & 0 \\
    \textbf{Spatial Decomp. \cite{jaderberg2014Spatial_decomp}} & 55.9 & 54.4 & 45.4 \\
    \textbf{DAC (Ours)} & \textbf{56.7} & \textbf{55.6} & \textbf{52.5}\\
    \hline
  \end{tabular}
  \end{adjustbox}
  \label{table:openpose_acc_comparison}
  \vspace{-2mm}
\end{table} 

Table \ref{table:openpose_acc_comparison} shows the accuracy of the model (after saving 40\%, 50\%, and 60\% FLOPs respectively) on COCO 2017 keypoint challenge. The parameters of Channel Decomposition and Spatial Decomposition are computed using Equation \ref{equation:rank_channel} and \ref{equation:rank_spatial} correspondingly. Compared to Channel and Spatial Decomposition, DAC achieves higher accuracy even when a significant amount of FLOPs is reduced. After saving 60\% FLOPs, Channel Decomposition cannot correctly detect any person's pose, while DAC can still achieve 7.1\% higher accuracy than Spatial Decomposition. 

Figure \ref{fig:openpose_vis} shows the visualized multi-person pose estimation results on COCO dataset. It shows that after being decomposed using DAC, the model still works pretty well. There are only small changes observed. For example, the decomposed model misses a leg of a person in the first example (the second person on the right side) and the third sample ( the second person on the left side). Please refer to our Appendix for more visualized results.

\subsection{Object Detection}
\begin{figure}[h]
	\vspace{-3mm}
	\centering
	\includegraphics[width=0.4\textwidth]{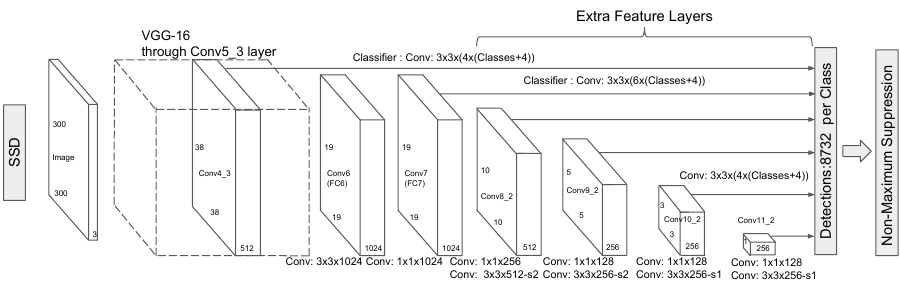}
	\caption{SSD architecture extracted from \cite{liu2016ssd}}
	\label{fig:ssd}
	\vspace{-5mm}
\end{figure}

Next, we evaluate the performance of DAC in the task of object detection using the Single Shot MultiBox Detector (SSD) model proposed by Liu et al. \cite{liu2016ssd}. 
%The SSD method produces a fixed-size collection of bounding boxes and scores for the presence of object class instances in these boxes using a convolutional network. The bounding boxes and scores are fed to a non-maximum suppression step to generate the final detections. 
Figure \ref{fig:ssd} shows the framework of the SSD.  

We use a model \footnote{https://github.com/pierluigiferrari/ssd\_keras} pre-trained on Pascal VOC2007 and VOC2012 trainval subset. The model uses VGG-16 \cite{simonyan2015very} as its base net that has (300x300) input size. Ten extra convolutional layers are added to the VGG-16 model to provide extra information. In total, 18 (3x3) convolutional layers and 5 (1x1) convolutional layers are used to generate multi-scale feature maps for detection, and 12 (3x3) convolutional layers are used to produce a fixed set of detection predictions. This model achieves 76.5\% on VOC2007 testing set. 

There is no benefit in decomposing a convolutional layer with (1x1) kernel size, so we only decompose those layers with (3x3) kernel size. Furthermore, considering that decomposing first layers causes large drops of accuracy, we do not decompose the first convolutional layer of the model. To simplify the description, we denote 18 layers (the first layer, conv1\_1, is not decomposed) that generate multi-scale feature maps by ``Feature Convolutional Layers (FL)'' and 12 layers that produce detection predictions by ``Detector Convolutional Layers (DL)''. 

\begin{figure}[h]
	\vspace{-3mm}
	\centering
	\includegraphics[width=0.4\textwidth]{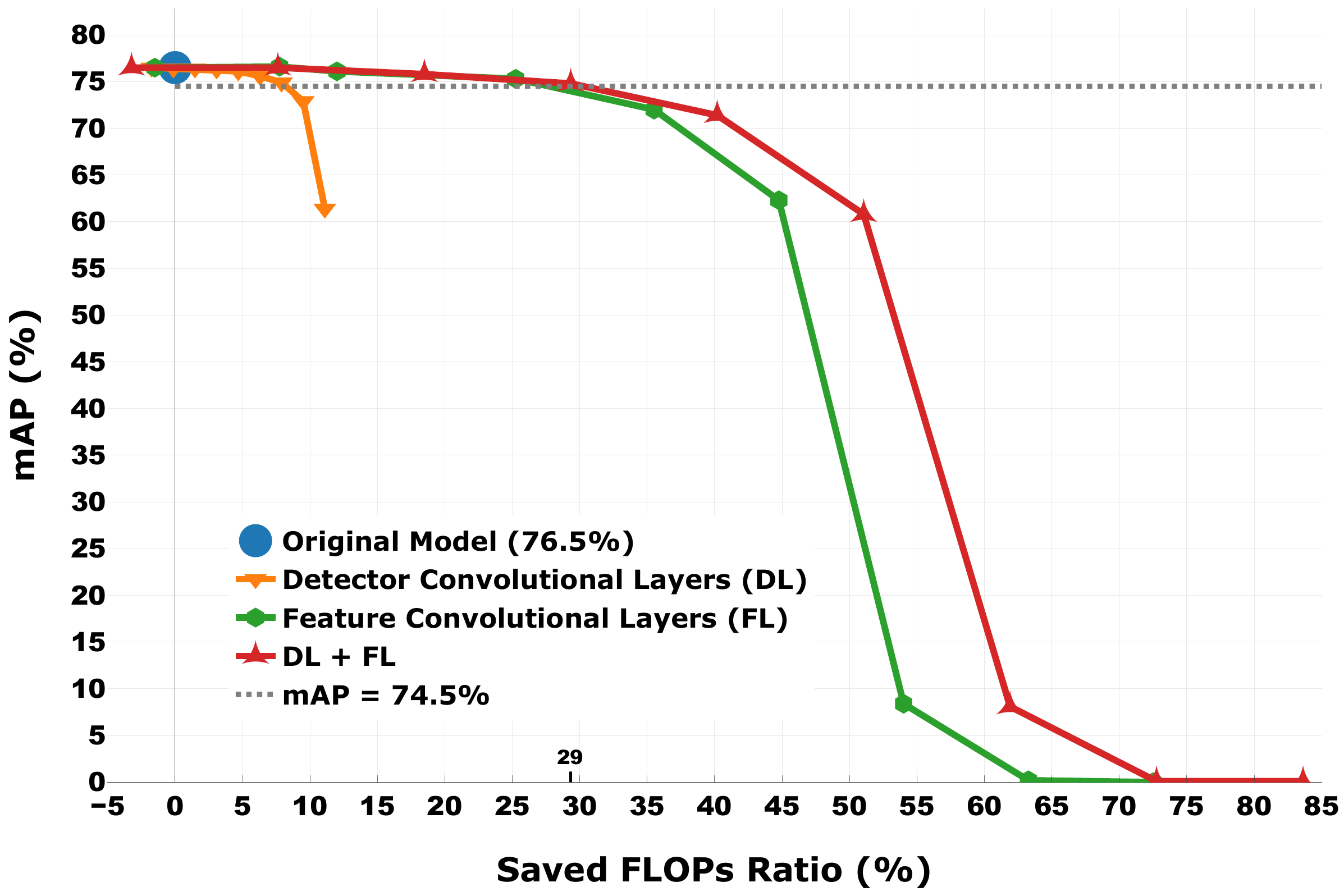}
	\vspace{-2mm}
	\caption{Object detection results on PASCAL VOC2007 testing set. Nine spots on each curve indicate Rank9 toward Rank1 correspondingly from left to right.}
	\label{fig:ssd_acc}
	\vspace{-3mm}
\end{figure}

\begin{figure*}[h]
	\centering
	\vspace{-2mm}
	\includegraphics[width=0.9\textwidth]{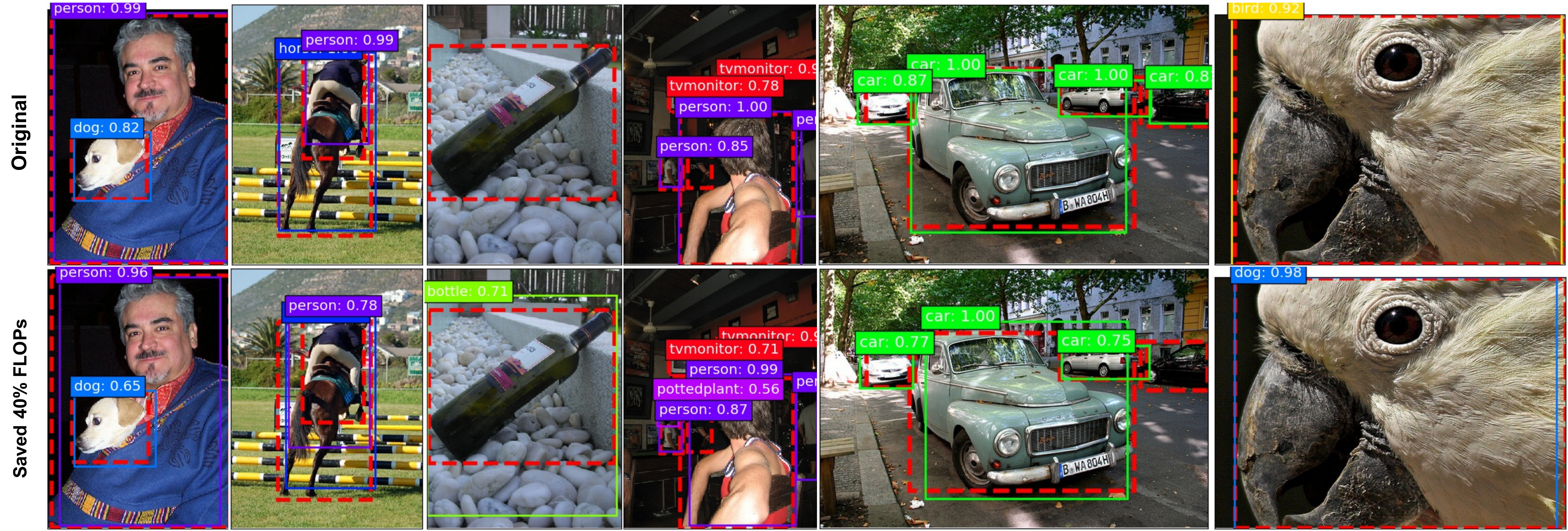}
	\caption{Visualized results on PASCAL VOC2007 dataset. The first row shows the results generated using the original weights, while the second row shows the results created using a model that saves 40\% FLOPs. Red dashed rectangles are ground truths. The first two samples are examples that the model works well after being decomposed, the third sample shows an example that DAC helps improve the performance, while the following three samples are different kinds of errors caused by decomposition.}
	\label{fig:ssd_vis}
	\vspace{-2mm}
\end{figure*}

\begin{table*}[h]
  \centering  
  \caption{Detection results on PASCAL VOC2007 testing set. All results expected original are collected using the SSD300 model decomposed with Rank6. ``DL'' indicates only Detector Convolutional Layers are decomposed and ``FL'' indicates only Feature Convolutional Layers are decomposed. }
  \begin{adjustbox}{width=\textwidth}
  \begin{tabular}{|l|c|c c c c c c c c c c c c c c c c c c c c|} 
    \hline
    Model & mAP(\%) & aero & bike & bird & boat & bottle & bus & car & cat & chair & cow & table & dog & horse & mbike & person & plant & sheep & sofa & train & tv \\
    \hline
    SSD\cite{liu2016ssd}(Original) & 76.5 & 78.6 & 83.9 & 75.3 & 67.8 & 48.5 & 86.7 & 84.7 & 87.7 & 58.1 & 79.3 & 75.0 & 85.9 & 87.5 & 82.6 & 77.5 & 51.2 & 77.1 & 79.5 & 87.2 & 76.5\\
    \hline
    (DL) Channel\cite{Jian2016channel-decomp} & 76.1 & 77.6 & 83.8 & 75.8 & 66.6 & 45.7 & 86.4 & 84.5 & 87.6 & 58.1 & 78.6 & 74.5 & 86.1 & 87.4 & 82.5 & 77.0 & 50.3 & 76.8 & 79.7 & 87.8 & 75.1 \\	    
    (DL) Spatial\cite{jaderberg2014Spatial_decomp} & 76.1 & 77.9 & 83.2 & 75.3 & 67.2 & 46.0 & 86.3 & 84.4 & 86.9 & 58.2 & 78.9 & 74.5 & 85.5 & 87.3 & 82.7 & 76.9 & 51.0 & 76.5 & 79.4 & 87.7 & 75.7\\
    \textbf{(DL) DAC(Ours)} & \textbf{76.3} & 78.4 & 82.9 & 74.5 & 68.3 & 47.8 & 86.7 & 84.4 & 88.4 & 58.0 & 79.4 & 74.9 & 85.6 & 86.5 & 83.1 & 77.3 & 50.7 & 77.3 & 79.0 & 87.5 & 76.1 \\
    \hline
    (FL) Channel\cite{Jian2016channel-decomp} & 62.2 & 70.4 & 69.7 & 63.8 & 52.9 & 38.3 & 75.1 & 79.8 & 72.8 & 42.2 & 73.2 & 38.0 & 65.7 & 76.3 & 69.6 & 64.0 & 38.8 & 66.6 & 53.4 & 75.6 & 57.3 \\
    (FL) Spatial\cite{jaderberg2014Spatial_decomp} & 63.2 & 73.7 & 69.7 & 64.6 & 52.0 & 39.0 & 75.6 & 79.9 & 77.6 & 42.6 & 73.2 & 39.5 & 70.7 & 76.3 & 71.3 & 65.5 & 37.9 & 67.0 & 53.0 & 77.9 & 56.1\\
    \textbf{(FL) DAC(Ours)} & \textbf{75.3} & 78.2 & 83.0 & 73.0 & 67.1 & 44.3 & 86.3 & 83.3 & 87.7 & 56.6 & 78.5 & 75.2 & 84.2 & 85.9 & 82.8 & 75.8 & 48.8 & 75.3 & 78.6 & 86.4 & 75.6\\
  	\hline
    (DL+FL) Channel\cite{Jian2016channel-decomp} & 62.2 & 70.6 & 69.4 & 64.1 & 51.1 & 36.1 & 75.7 & 79.8 & 72.8 & 43.0 & 72.9 & 39.9 & 66.4 & 74.5 & 70.2 & 63.7 & 38.5 & 65.9 & 55.1 & 75.4 & 58.7 \\
	  (DL+FL) Spatial\cite{jaderberg2014Spatial_decomp} & 63.1 & 73.8 & 70.3 & 64.1 & 50.8 & 37.8 & 75.3 & 79.8 & 76.9 & 43.0 & 74.3 & 39.8 & 69.7 & 75.8 & 70.9 & 64.5 & 38.6 & 68.9 & 54.1 & 77.9 & 56.2 \\
    \textbf{(DL+FL) DAC(Ours)} & \textbf{74.8} & 76.4 & 81.1 & 73.1 & 66.0 & 44.6 & 85.9 & 83.1 & 88.1 & 56.5 & 76.8 & 74.0 & 84.3 & 86.1 & 83.1 & 75.5 & 47.7 & 74.1 & 77.5 & 86.1 & 75.7 \\
    \hline
  \end{tabular}
  \end{adjustbox}
  \label{table:ssd_acc_detail}
  \vspace{-3mm}
\end{table*} 

\begin{table}[h]
  \centering  
  \caption{Object detection results on PASCAL VOC2007 Dataset.}
  \begin{adjustbox}{width=0.45\textwidth}
  \begin{tabular}{|l|c c c|} 
    \cline{2-4} 
  \multicolumn{1}{c|}{}& \multicolumn{3}{|c|}{\textbf{Mean Average Precision (\%)}}  \\
  \hline
  \multicolumn{1}{|c|}{\textbf{SSD \cite{liu2016ssd} (Original) }} & \multicolumn{3}{|c|}{76.5} \\
    \hline
    \hline
    \textbf{Method} & \textbf{Saved 30\%} & \textbf{Saved 40\%} & \textbf{Saved 50\%} \\
    \hline    
	\textbf{Channel Decomp. \cite{Jian2016channel-decomp}} & 62.2 & 60.0 & 52.4 \\
    \textbf{Spatial Decomp. \cite{jaderberg2014Spatial_decomp}} & 63.1 & 62.2 & 60.6 \\
    \textbf{DAC (Ours)} & \textbf{74.8} & \textbf{71.4} & \textbf{60.8}\\
    \hline
  \end{tabular}
  \end{adjustbox}
  \label{table:ssd_acc_comparison}
  \vspace{-5mm}
\end{table} 

We demonstrate the experimental results in Figure \ref{fig:ssd_acc}. First, one can see that if 2\% mAP drop is acceptable, DAC saves up to 29\% FLOPs. Second, decreasing the decomposition rank results in a drop of mAP, which is also observed in the previous experiment. Third, compared to ``DL'', ``FL'' achieves a bigger FLOPs saved ratio. This is because that there are fewer layers in ``DL'' and each layer in ``DL'' has fewer channels than layers in ``FL''. In addition, for this model, the maximum decomposition rank is 9 so when the decomposition rank is set to 9, the number of parameters increases after decomposition. This is because that all layers we decompose in this model have (3x3) kernel size whose full rank is 9. The newly generated depthwise layer with Rank9 has the same number of parameters as the decomposed layer, while an extra pointwise layer that has $rC \times N \times 1 \times 1$ parameters is added. 
%Because full rank can duplicate the original function the accuracies do not reduce when full rank (Rank9) is used.

Table \ref{table:ssd_acc_comparison} shows the comparison of the detection accuracy on PASCAL VOC2007 Dataset. One can see that DAC achieves higher accuracy than other schemes. In Table \ref{table:ssd_acc_detail}, we list the details of the detection results on PASCAL VOC2007 testing set. Comparing the results of DAC to the original model, one can see that decomposing the model using DAC does not impact the performance of the model too much, for all categories. The change of the accuracy happens on each category within a small range. 

Figure \ref{fig:ssd_vis} shows the visualized object detection results on PASCAL VOC2007 testing set. From the first two samples, one can see that after being decomposed, the model can still correctly detect objects. The locations and sizes of the detected bounding boxes have small changes. The third sample is an example that the original model does not detect an object (the bottle) that is successfully detected by our decomposed model. The fourth sample shows an extra false positive example (an unexpected potted-plant is detected), the fifth sample is a missing example (miss the car on the right), and the last sample is an example that the detected label changed (from bird to dog). Please refer to our Appendix for more visualized results.

%\subsection{Hardware Friendly}

%% file: conclusion.tex
%-------------------------------------------------------------------
%\newpage
\section{Conclusion}
\label{conclusion}

In this paper, we propose a novel decomposition method, namely DAC. 
Given a pre-trained model, DAC is able to factorize an ordinary convolutional layer into two layers with much fewer parameters and computes their weights by decomposing the original weights directly. Thus, no training (or fine-tuning) or any data is needed. The experimental results on three computer vision tasks show that DAC reduces a large number of FLOPs while maintaining high accuracy of a pre-trained model.

We plan to evaluate the performance of DAC for deep learning models in other fields, e.g., voice recognition, language translation, etc. We also want to explore the possibility of adapting DAC on other types of layers, e.g. 3D convolutional layer, compared with other tensor decomposition formats \cite{kim2015Tucker_decomp,lebedev2014CP_decomp}. Another research direction is to combine low rank constraints with weight decomposition. These constraints could be convex regularizations like nuclear norm and Frobenius norm, or non-convex quasi-norms like Schatten $p$ and TS1 \cite{shuai-DCATL12018,shuai-TL12017,shuai-TS12017}. 

%% file: acknowledgements.tex
\section{Acknowledgements}
We would love to express our appreciation to Jacob Nelson for his useful discussions.